\documentclass{article}
\usepackage{ijcai20}

\usepackage{times}
\usepackage{soul}
\usepackage{url}
\usepackage[hidelinks]{hyperref}
\usepackage[utf8]{inputenc}
\usepackage[small]{caption}
\usepackage{graphicx}
\usepackage{amsmath}
\usepackage{amsthm}
\usepackage{booktabs}
\usepackage{algorithm}
\usepackage{algorithmic}

\usepackage{subfigure}
 
\usepackage{dblfloatfix}
\usepackage{tasks}

\newcommand{\eg}{\textit{e}.\textit{g}. }

\title{A 3D Convolutional Approach to Spectral Object Segmentation in Space and Time}
\author{
Elena Burceanu$^{1,2}$\footnote{Contact Author}\and
Marius Leordeanu$^{3, 4}$\\
\affiliations
$^1$Bitdefender\\
$^2$University of Bucharest, Romania\\
$^3$Institute of Mathematics of the Romanian Academy\\
$^4$University Politehnica of Bucharest, Romania\\
\emails
eburceanu@bitdefender.com,
marius.leordeanu@cs.pub.ro
}

\begin{document}

\maketitle

\begin{abstract}
We formulate object segmentation in video as a spectral graph clustering problem in space and time, in which nodes are pixels and their relations form local neighbourhoods. We claim that the strongest cluster in this pixel-level graph represents the salient object segmentation. We compute the main cluster using a novel and fast 3D filtering technique that finds the spectral clustering solution, namely the principal eigenvector of the graph's adjacency matrix, without building the matrix explicitly - which would be intractable. Our method is based on the power iteration which we prove is equivalent to performing a specific set of 3D convolutions in the space-time feature volume. This allows to avoid creating the matrix and have a fast parallel implementation on GPU. We show that our method is much faster than classical power iteration applied directly on the adjacency matrix. Different from other works, ours is dedicated to preserving object consistency in space and time at the level of pixels. 
In experiments, we obtain consistent improvement over the top state of the art methods on DAVIS-2016 dataset. We also achieve top results on the well-known SegTrackv2 dataset.
\end{abstract}

\section{Introduction}
\begin{figure}[h]
	\begin{center}
		\includegraphics[width=0.85\linewidth]{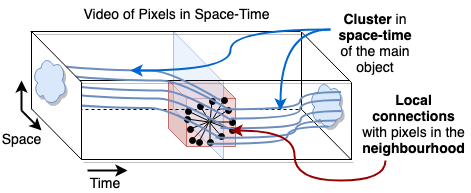}
	\end{center}
	\caption{We see the video as a locally connected graph of pixels in space-time. The strength and the number of connections are enforcing the pixel membership to the salient video object.}
	\label{fig: intuition}
\end{figure}

Elements from a video are interconnected in space and time and have an intrinsic graph structure (Fig.~\ref{fig: intuition}). Most existing approaches use higher-level components, such as objects, super-pixels or features, at a significantly lower resolution. Considering this graph structure in space-time, explicitly at the dense pixel-level, is an extremely expensive problem. Our proposed solution to video object segmentation, Spectral Filtering Segmentation (\textbf{SFSeg}), is based on transforming an expensive eigenvalue problem inspired from spectral clustering, into 3D convolutions on the space-time volume. This makes it fast, while keeping the properties of spectral clustering. We are the first, to our best knowledge, to propose a practical spectral clustering approach to video object segmentation at the pixel level, in space and time.

Most state of the art algorithms for this task do not use the time constraint, and when they do, they take little advantage of it. Time plays a fundamental factor in how objects move and change in the world, but computer vision does not yet exploit it sufficiently. Consequently, the segmentation outputs of current state of the art algorithms is not always consistent over time. Our work comes to address precisely this aspect and our contribution is demonstrated through solid experiments on DAVIS-2016 and SegTrackv2 datasets on which we improve over state of the art methods.

We demonstrate in experiments that the eigenvector of the graph's adjacency matrix is a good solution for salient object segmentation. Once our filtering-based optimization converges, the segmentation map is spatio-temporally consistent, with a smooth transition between frames: noise coming from other objects is removed and missing parts of the object are added back. Through multiple iterations, the relevant information is propagated step by step to farther away neighbourhoods in space and time, acting like a diffusion.

\textbf{Our contribution} is two-fold. Besides formulating the segmentation problem in video as an eigenvalue problem on the adjacency matrix of the graph in space-time, we also provide a very fast optimization algorithm that computes the required eigenvector (which represents the desired segmentation) without explicitly creating or using the huge adjacency matrix. We prove theoretically and in practice that our algorithm reaches the same solution as a standard routine for eigenvector computation. We also show in experiments that the values in the final eigenvector, with one element per video pixel, confirm the spectral clustering assumption and provide an improved soft-segmentation of the main object. 

\section{Related work}
\label{sec: related_work}

Most state of the art methods for video object segmentation are using \textbf{CNNs architectures}, pre-trained for object segmentation on other large image datasets. They have a strong image-based backbone and are not designed from scratch with both space and time dimensions in mind. Many solutions \cite{masktrack} adapt image segmentation methods by adding an additional branch to the architecture for incorporating the time axis: motion branch (previous frames or optical flow as) or previous masks branch (for mask propagation). Other methods are based on one-shot learning strategies and fine tune the model on the first video frame, followed by some post-processing refinement \cite{osvoss}. Approaches derived from OSVOS~\shortcite{osvos} do not take the time axis into account. Our method comes to better address the natural space-time relationship, which is why it is effective when combined with frame-based segmentation algorithms. 

\paragraph{Graph representations.} Graph methods are suitable for segmentation and can have different representations, where the \textbf{nodes} can be pixels, super-pixels, voxels or image/video regions. Graph edges are usually undirected, modeled as symmetric similarity functions.
The choice of the representation influences both accuracy and runtime. Specifically, pixel-level representations are computationally extremely expensive, making the problem intractable for high resolution videos. Our fast solution implicitly uses a pixel-level graph representation: we make
a first-order Taylor approximation of the Gaussian kernel (usually used for pairwise affinities) and rewrite it as a sequence of 3D convolutions in the video directly. Thus, we get the desired outcome without explicitly working with the graph. We describe it in detail in Sec.~\ref{sec: math_formulation}. 

\paragraph{Spectral clustering.} Computing eigenvectors of matrices extracted from data is a classic approach for clustering. There are several choices in the literature for choosing those matrices, the most popular being the Laplacian matrix \cite{NJW}, normalized \cite{img_normalized_cut_malik_2000} or unnormalized. Other methods use the random walk matrix or directly the unnormalized adjacency matrix. Most methods are based on finding the eigenvectors corresponding to the smallest eigenvalues, while others, including our approach, require the leading eigenvectors. Graph Cuts are a popular class of spectral clustering algorithms, with many variants: normalized, average, min-max, mean cut and topological cut. 

\paragraph{CRFs.} Discriminative graphical models~\cite{kumar2003discriminative} 
are often applied over the segmentation of images and videos (denseCRF~\cite{denseCRF}). CRFs are effective as they incorporate the observed data both at the level of nodes as well as edges. But they have a strict probabilistic interpretation and use inference algorithms that are significantly more expensive than the simpler eigenvector power iteration that we use for optimizing our non-probabilistic objective score.

\paragraph{Image segmentation.} Graph cuts have been used in image segmentation \cite{img_normalized_cut_malik_2000}. They are expensive in practice, as they require the computation of eigenvectors of smallest eigenvalues for very large Laplacian matrices. Fast graph-based algorithm for image segmentation exist, such as \cite{img_efficient_graph_based_felzen_2004}, which is linear in the number of edges and it is based on an heuristic for building the minimum spanning tree. It is still used as staring point by current methods. Another approach~\cite{img_weakly_superv_2015} is to learn image regions with spectral graph partitioning and formulate segmentation as a convex optimization problem.

\paragraph{Video Segmentation.} Many video segmentation methods adapt existing image segmentation. In \cite{video_efficient_param_graph_part_2015} a parametric graph partitioning model over superpixels is proposed. Hierarchical graph-based segmentation over RGBD video sequences \cite{video_efficient_hierarch_rgbd_2018} also groups pixels into regions. The problem is solved using bipartite graph matching and minimizing the spanning tree. In \cite{video_unsup_segm_motion_2018}, an efficient graph cut method is applied on a subset of pixels. To our best knowledge, all of the efficient methods group pixels into superpixels, regions from a grid or object proposals to handle the computational and memory burden. However, the hard initial grouping of pixels comes with a risk and could carry errors into the final solution, as it misses details available only at the original pixel resolution.

\textbf{Our formulation} is most related to \cite{marius_iccv2005,meila_shi}. Our solution is the leading eigenvector of $\mathbf{M}$ (the adjacency matrix), computed fast and stably with power iteration as explained in Sec.~\ref{sec: math_formulation}. Note that using the unnormalized adjacency matrix in combination with power iteration is the least expensive spectral approach and the only one that can be factored into simple and fast 3D convolutions. This possibility gives our algorithm efficiency and speed (Sec.~\ref{sec: algorithm}).

\section{Our approach}
\label{sec: math_formulation}

We formulate salient object segmentation in video as a graph partitioning problem (foreground vs background), where the graph is both spatial and temporal. Each node $i$ represents a pixel in the space-time volume, which has $N = N_f \times H \times W$ pixels. $N_f$ is the number of frames and $(H,W)$ the frame size. Each edge captures the similarity between two pixels and is defined by the pairwise function $\mathbf{M}_{i, j}$. The pairwise connections between pixels $i$ and $j$, in space and time are symmetric and always non-negative, defining a $N \times N$ adjacency matrix $\mathbf{M}$. We take into account only the local connections in space-time, so $\mathbf{M}$ is sparse. 

Let $\mathbf{s}$ and $\mathbf{f}$ be feature vectors of size $N \times 1$ with a feature value for each node. They will be used in defining the similarity function $\mathbf{M}_{ij}$ (Eq.~\ref{eq: mij_equation}).
For now we consider the simplest case when $(\mathbf{s}_i,\mathbf{f}_i)$ represent single channel features (e.g. they could be soft masks, grey level values, edge or motion cues, or any pre-trained features). Later on we show how we can easily adapt the formulation to the multi-channel feature case. We define the edge similarity $\mathbf{M_{i, j}}$ using a Gaussian kernel:

\begin{equation}
    \begin{aligned}
        \mathbf{M}_{i,j} &= \mathbf{s}_i^p \mathbf{s}_j^p e^{- \alpha ( \mathbf{f}_i -  \mathbf{f}_j)^2 - \beta \mathbf{dist}^2_{i, j}} \\
        &= \mathbf{s}_i^p \mathbf{s}_j^p e^{- \alpha ( \mathbf{f}_i -  \mathbf{f}_j)^2} \mathbf{G}_{i, j}
    \end{aligned}
	\label{eq: mij_equation}
\end{equation}

\begin{equation}
    \begin{aligned}
    	 \mathbf{M}_{i,j} 
    	 & \approx \underbrace{\mathbf{s}_i^p \mathbf{s}_j^p}_{\textnormal{unary terms}} \underbrace{[1 - \alpha (\mathbf{f}_i -  \mathbf{f}_j)^2] \mathbf{G}_{i, j}}_{\textnormal{pairwise terms}}.
    \end{aligned}
	\label{eq: mij_equation_approx}
\end{equation}
In graph methods, it is common to use two types of terms for representing the model over the graph. Unary terms are about individual node properties, while pairwise terms describe relations between pairs of nodes. In our case, $\mathbf{s}_i$, $\mathbf{s}_j$ describe individual node properties, whereas $\mathbf{f}_i$, $\mathbf{f}_j$ are used to define the pairwise similarity kernel between the two nodes. Note that in Eq.~\ref{eq: mij_equation_approx} we approximate the Gaussian kernel with its first-order Taylor expansion. The approximation is crucial in making our filtering approach possible, as shown next. Hyperparameters $p$ and $\alpha$ control the importance of those terms.

To partition the space-time graph of video pixels, we want to find the strongest cluster in this graph. We first represent a segmentation solution (i.e., cluster in the space-time graph) with an indicator vector $\mathbf{x}$, that has one element for each node in the 3D space-time volume, such that $x_i=1$ if node (pixel) $i$ is in the video segmentation cluster (foreground) and  $x_i=0$ otherwise (background). We define the clustering score to be the sum over all pairwise similarity terms $\mathbf{M}_{ij}$ between the nodes inside the cluster. The higher this score, the stronger the sum of connections and the cluster. The segmentation score can be written compactly in matrix form as $S(\mathbf{x})=\mathbf{x}^T\mathbf{M}\mathbf{x}$. Similar to other spectral approaches in graph matching \cite{marius_iccv2005}, we find the segmentation solution $\mathbf{x}_s$ that maximizes $S(\mathbf{x})$ under the relaxed constraints $\|\mathbf{x}\|_2 = 1$. Fixing the L2 norm of $\mathbf{x}$ is needed since only relative soft segmentation values matter. Thus, our optimization problem become one of maximizing the Raleigh quotient:

\begin{equation}
    \begin{aligned}
        \mathbf{x}_s = \underset{\mathbf{x}}{\mathrm{argmax}}  (\mathbf{x}^T \mathbf{M} \mathbf{x } / \|\mathbf{x}\|_2).
    \end{aligned}
	\label{eq: argmax_eq}
\end{equation}

The global optimum solution is the principal eigenvector of $\mathbf{M}$. $\mathbf{M}$ is symmetric and has non-negative values, so the solution will also have non-negative elements, by Perron-Frobenius theorem \cite{frobenius}. The final segmentation could be simply obtained by thresholding. However, matrix $\mathbf{M}$, even for a small video has 20 million nodes (50 frames of $480 \times 854$), making the problem of finding the leading eigenvector with standard procedures intractable (Sec~\ref{subsec: numerical_complexity}).

Next we show how to take advantage of the first-order expansion of the pairwise terms defining $\mathbf{M}$ and break power iteration into several very fast 3D convolutions in space and time, directly on the feature maps, without explicitly using the very big adjacency matrix. Our method receives as input pixel level feature maps and returns a final segmentation, as the solution $\mathbf{x}_s$ to problem \ref{eq: argmax_eq}.

\subsection{Power iteration with pixel-wise iterations}
We apply power iteration algorithm to compute the eigenvector. At iteration $k+1$, we have Eq.~\ref{eq: power_it_eq}:
\begin{equation}
	\mathbf{x}^{k+1}_i \leftarrow \sum_{j \in \mathcal{N}(i)}\mathbf{M}_{i, j} \mathbf{x}^k_j, 
	\label{eq: power_it_eq}
\end{equation}
where, after each iteration, the solution is normalized to unit norm and $\mathcal{N}(i)$ is the set of neighbors pixels with $i$, in space and time. Expanding $\mathbf{M}_{i, j}$ (Eq.~\ref{eq: mij_equation_approx}), Eq.~\ref{eq: power_it_eq} becomes:

\begin{equation}
\mathbf{x}^{k+1}_i \leftarrow \alpha \mathbf{s}_i^p \sum_{j \in \mathcal{N}(i)} \mathbf{s}_j^p [\alpha^{-1} - \mathbf{f}_i^2 - \mathbf{f}_j^2 + 2 \mathbf{f}_i \mathbf{f}_j] \mathbf{G}_{i, j} \mathbf{x}^k_j,
\end{equation}

\begin{equation}
\label{eq: before_matrix_form}
  \begin{aligned}
    \mathbf{x}^{k+1}_i \leftarrow  \alpha \mathbf{s}_i^p ( \alpha^{-1} - \mathbf{f}_i^2) \sum_{j \in \mathcal{N}(i)} \mathbf{s}_j^p  \mathbf{G}_{i, j} \mathbf{x}^k_j -\\
    \alpha \mathbf{s}_i^p \sum_{j \in \mathcal{N}(i)}  \mathbf{s}_j^p \mathbf{f}_j^2 \mathbf{G}_{i, j} \mathbf{x}^k_j + \\
    2 \alpha \mathbf{s}_i^p \mathbf{f}_i  \sum_{j \in \mathcal{N}(i)} \mathbf{s}_j^p \mathbf{f}_j \mathbf{G}_{i, j} \mathbf{x}^k_j.
  \end{aligned}
\end{equation}

\subsection{Power iteration using 3D convolutions}
In Eq.~\ref{eq: before_matrix_form} we observe that the links between the nodes are local ($\textbf{M}$ is sparse) and we can replace the sums over neighbours with local 3D convolutions in space and time. Thus, we rewrite Eq.~\ref{eq: before_matrix_form} as a sum of convolutions in 3D:

\begin{equation}
\begin{aligned}
\mathbf{X}_{crt} \leftarrow \mathbf{S}^p \cdot (\alpha^{-1} \mathbf{1} - \mathbf{F}^2) \cdot G_{3D} * (\mathbf{S}^p  \cdot \mathbf{X}^k) - \\
                        \mathbf{S}^p \cdot G_{3D} * (\mathbf{F}^2 \cdot \mathbf{S}^p \cdot \mathbf{X}^k) + \\
                        2 \mathbf{S}^p \cdot \mathbf{F}\cdot G_{3D} * (\mathbf{F} \cdot \mathbf{S}^p \cdot \mathbf{X}^k),
\end{aligned}
\label{eq: matrix_full_eigenvector}
\end{equation}

\begin{equation}
\mathbf{X}^{k+1} \leftarrow \mathbf{X}_{crt} / \|\mathbf{X}_{crt}\|_2,
\end{equation}
where $*$ is a convolution over a 3D space-time volume with a 3D Gaussian filter ($G_{3D}$), $\cdot$ is an element-wise multiplication, 3D matrices $\mathbf{X}^k, \mathbf{S}, \mathbf{F}$ have the original video shape ($N_f \times H \times W$) and $\mathbf{1}$ is a 3D matrix with all values 1. We transformed the 
standard form of power iteration in Eq.~\ref{eq: power_it_eq} in several very fast matrix operations: 3 convolutions and 13 element-wise matrix operations (multiplications and additions), which are local operations that can be parallelized. 
\subsection{Multiple feature channels}
Our approach in Eq.~\ref{eq: matrix_full_eigenvector} can easily accommodate multiple feature channels if we rewrite $\mathbf{M}_{i, j}$ from Eq.~\ref{eq: mij_equation_approx} and propagate it through Eq.~\ref{eq: matrix_full_eigenvector}, the final multi-channel solution is obtained by summing over the final solution for each channel:


\begin{equation}
\label{eq: multi_channel}
    \begin{aligned}
        \mathbf{X}_{crt}^{multi} = \sum_{m = 1}^{N_{feat}} \mathbf{X}_{crt} (\mathbf{F}_{c}),
    \end{aligned}
\end{equation}
where $\mathbf{F}_{c}$ is one (3D) channel feature matrix.

\section{Algorithm}
\label{sec: algorithm}
We present the version of our algorithm (Alg.~\ref{alg: power_iteration}) that has a single channel feature map, but can be easily adapted to the multi-channel version, using Eq.~\ref{eq: multi_channel}. We first initialize the solution $\mathbf{X}$ with a uniform vector or with a soft-segmentation provided by another method, if it is available. We also initialize feature maps
$\mathbf{S}$ and $\mathbf{F}$, which could be of any kind: lower-level (optical flow, edges, gray-level values) or higher-level pre-trained semantic features (deep features or initial soft/hard segmentation maps). 
At each iteration, we select a time frame around the current one. In Step 2, we multiply the corresponding matrices, apply the convolutions, compose the results and obtain the new segmentation mask for pixels in current frame, using the space-time operations (as in Eq.~\ref{eq: matrix_full_eigenvector}). Since the solution needs to be binary at the end (for evaluation), after each iteration (Step 3, line 14 in Alg.~\ref{alg: power_iteration}), we project the solution in a more discrete space (see Sec.~\ref{subsec: binarization}). 

\begin{algorithm}[h]
    \caption{Power iteration with 3D convolutions algorithm.  At each iteration we pass through the whole video and compute the updated soft-segmentation $X$. First, we  select a time window around current frame $[i-w, i+w]$. Secondly we compute the eigenvector with convolutions. Then, after each iteration, we binarize the solution (see Sec.~\ref{subsec: binarization}).}
    \label{alg: power_iteration}
    $\mathbf{S}$ - unary feature maps for video\\
    $\mathbf{F}$ - defines pairwise feature maps for video\\
    $\mathbf{X}$ - salient object segmentation 
    \begin{algorithmic}[1]
        \STATE $ \mathbf{X} \leftarrow \mathbf{S}$\\
        \FOR{$iter$ {\bfseries in} $[1.. N_i]$} {
            \FOR{$i$ {\bfseries in} $[1.. N_f]$} {
                \STATE \(\triangleright\) STEP 1. Take a temporal window around frame $i$:\\
                \STATE $\mathbf{S}_w, \mathbf{X}_w, \mathbf{F}_w \leftarrow T_{OF}(\mathbf{S}, \mathbf{X}, \mathbf{F}) [i-w:i+w]$\\
                \STATE \(\triangleright\) STEP 2. Compute new mask:\\
                \STATE $\mathbf{T1} \leftarrow  (\alpha^{-1}\mathbf{1} - \mathbf{F}_w^2) \cdot G_{3D} * (\mathbf{S}^p_w \cdot \mathbf{X}_w) $\\
                \STATE $\mathbf{T2} \leftarrow - G_{3D} * (\mathbf{F}_w^2 \cdot \mathbf{S}_w^p \cdot \mathbf{X}_w)$\\
                \STATE $\mathbf{T3} \leftarrow 2 \mathbf{F}_w \cdot G_{3D} * (\mathbf{F}_w \cdot \mathbf{S}_w^p \cdot \mathbf{X}_w)$\\
                \STATE $\mathbf{X}_{new}[i] \leftarrow \mathbf{S}_w^p \cdot (\mathbf{T1} + \mathbf{T2} + \mathbf{T3}) $\\
            }
            \ENDFOR
            \STATE $\mathbf{X} \leftarrow$ normalize$(\mathbf{X}_{new})$\\
            \STATE \(\triangleright\) STEP 3. binarization:\\
            \STATE $\mathbf{X} \leftarrow$ project\_binary$(\mathbf{X})$
        }
        \ENDFOR
    \end{algorithmic}
\end{algorithm}

\subsection{Binarization - Spectral vs Discrete space}
\label{subsec: binarization}
At the end, we need to have a hard segmentation map for the object of interest. Over the iterations, a spectral method makes the solution continuous. It was previously observed that in graph matching optimization, where the solution is relaxed \cite{ipfp}, keeping it close to the initial discrete domain comes with a better final performance, even though the optimum in the spectral space is affected. So we integrated the binarization in the iterative optimization.
After a few iterations in the continuous space, we start projecting the solution on an almost discrete space through a sigmoid (which continuously approximates a step function) and initialize the next iteration with this projection. After the last iteration, we apply a hard threshold on a solution much closer to the discrete space than before. This way, the transition is smoother compared with a simple sharp thresholding.

\subsection{Numerical Analysis}
\label{subsec: numerical_complexity}
We compare the standard power iteration eigenvector computation with our filtering formulation, both from qualitative and quantitative (speedup) points of view.

\paragraph{Computational Complexity.} Lanczos \shortcite{lanczos_1950} method for sparse matrices has $\mathcal{O}(k N_f N_p N_i)$ complexity for computing the leading eigenvector,
where $k$ is the number of neighbours for each node, $N_f$ the number of frames in video, $N_p$ the number of pixels per frame and $N_i$ the number of iterations. Our full iteration algorithm has also $\mathcal{O}(k N_f N_p N_i)$ complexity, but with highly parallelizable operations, comparing to Lanczos.  The Gaussian filters are separable, so the 3D convolutions can be broken into a sequence of three vector-wise convolutions, reducing the complexity $\mathcal{O}(k)$ for filtering to $3 \mathcal{O}(k^{\frac{1}{3}})$: $3$*$7$*$7$=$147$ vs $3$+$7$+$7$=$17$ for a $3$x$7$x$7$ kernel.

\begin{figure}[t]
	\begin{center}
		\includegraphics[width=0.9\linewidth]{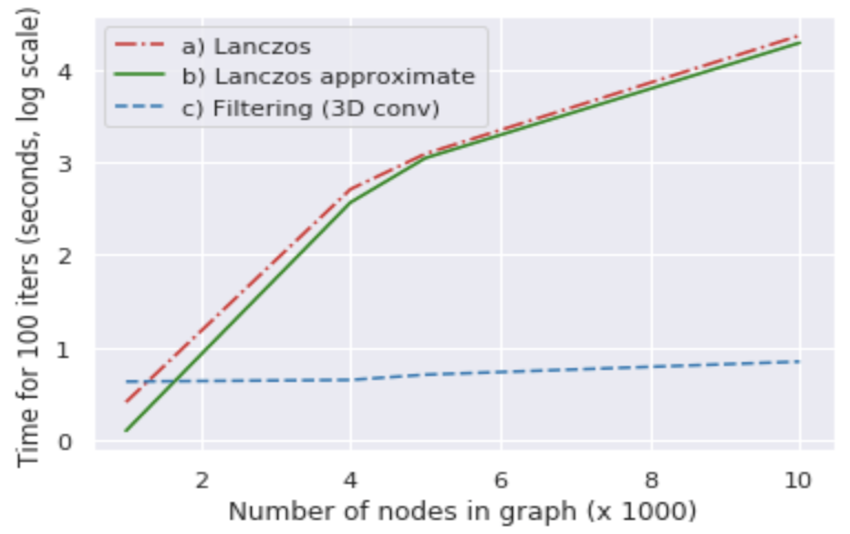}
	\end{center}
    \caption{Total runtime in logarithmic scale for 100 iterations, including the time for building the adjacency matrix for power iteration. Our filtering formulation scales with the number of nodes, in contrast to power iteration, having an exponentially better time.}
    \label{fig: time_pi_fi}
\end{figure}

\begin{figure}[h]
	\begin{center}
	    \includegraphics[width=0.85\linewidth]{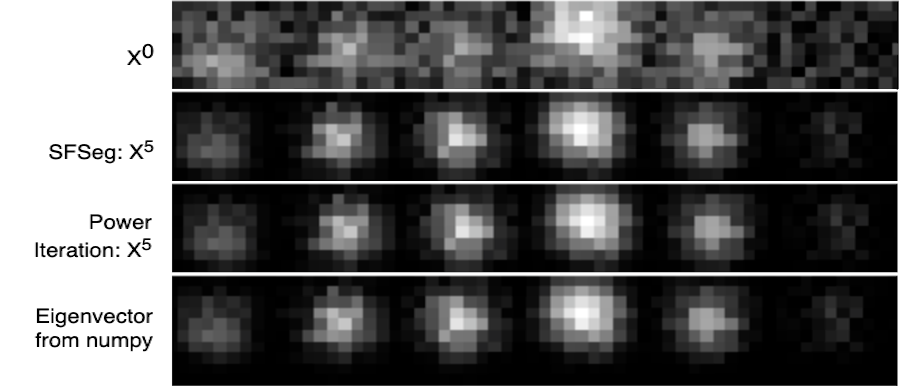}
	\end{center}
    \caption{A toy examples for qualitative comparisons with soft masks for a six frames video. Starting with a very noisy input segmentation mask and showing: SFSeg segmentations after 5 iters; Power Iteration after 5 iters; the real main eigenvector. The results are almost identical, proving that SFSeg is a good approximation. More, for other methods, this is tractable only on toy examples.}
    \label{fig: fi_iters}
\end{figure}

We compare three solutions: \textbf{a)} Lanczos for the principal eigenvector for Eq.~\ref{eq: mij_equation} \textbf{b)} Lanczos for the approximate adjacency matrix as in Eq.~\ref{eq: mij_equation_approx} \textbf{c)} our 3D convolutions approach. For a small graph of 4000 nodes (a video with 10 frames of $20 \times 20$ pixels), \textbf{a)} and \textbf{b)} have 0.15 sec/iter and our 3D filtering formulation has 0.02 sec/iter (Fig.~\ref{fig: time_pi_fi}). Our approach scales better, having a huge advantage when working with videos with millions of nodes because we do not explicitly build the adjacency matrix and filtering is parallelized on GPU.

\paragraph{Qualitative analysis.} We perform tests on synthetic data, in order to study the differences between the original spectral solution using the exponential pairwise scores (\ref{eq: mij_equation}) and the one obtained after our first-order Taylor approximation trick (\ref{eq: mij_equation_approx}). In Fig. \ref{fig: fi_iters} we see qualitative comparisons between the solutions obtained by three implementations: our SFSeg, power iteration with original pairwise scores and \textit{numpy} eigenvector with original pairwise scores. The output is almost identical. In the synthetic experiments, the input is noisy, but all spectral solutions manage to reconstruct the initial segmentation.

\begin{figure}[t]
\begin{center}
    \includegraphics[width=0.9\linewidth]{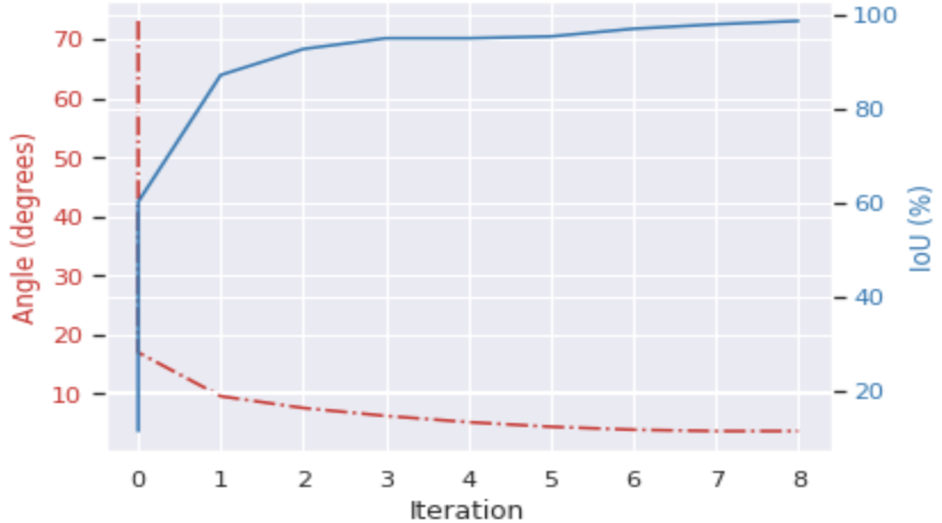}
    \end{center}
	\caption{The angle and the IoU between real eigenvector and our SFSeg solution. The evolution of those metrics is monitored over multiple SFSeg iterations.}
	\label{fig: approx}
\end{figure}

\paragraph{Quantitative analysis.} We analyze the numerical differences between the original eigenvector and our approximation (SFSeg). We plot the angle (in degrees) and the IoU (Jaccard) between SFSeg (first-order approximation of pairwise functions, optimized with 3D convolutions) and the original eigenvector (exponential pairwise functions in the adjacency matrix), over multiple SFSeg iterations in Fig.~\ref{fig: approx}. Note that in these experiments we intentionally start from a far away solution (70 degrees difference between the SFSeg initial segmentation vector and the original eigenvector) to better show that SFSeg indeed converges to practically the same eigenvector.
Such comparisons can be performed only on synthetic data with relatively small videos, for which the computation of the adjacency matrix needed for the original eigenvector is tractable. The results clearly show that SFSeg, with first order approximations of the pairwise functions on edges and optimization based on 3D filters, reaches the same theoretical solution, while being orders of magnitude faster.

\section{Experimental Analysis}
\label{sec: experiments}

\paragraph{Experiments on DAVIS-2016.} DAVIS-2016~\cite{davis2016} is a densely annotated video object segmentation dataset. It contains 50 high-resolution video sequences (30 train/20 valid), with a total of 3455 annotated frames of real-world scenes. The benchmark comes with two tasks: the unsupervised one, where the solutions do not have access to the first frame of the video and the semi-supervised one, where the methods use the ground-truth from the first frame. In both setups, the methods can train the model on the training set and report their performance on the validation set. Our results are reported on the validation set, but we do not use the training set. For optical flow we used the Pytorch implementation of Flownet2 \cite{flownet2-pytorch}.

\paragraph{Experimental Setup.} We test SFSeg with input from pre-computed segmentations of the video produced by top methods from DAVIS-2016, on both tasks. For the features maps, we initialized $\mathbf{S}$ with the pre-computed input segmentation values. For $\mathbf{F}$, we used two channels: the magnitude for the direct optical flow and for the reverse optical flow. We set: $N_{i} = 5$; $\alpha = 1$ and $p = 0.1$ for unsupervised task and $p = 0.2$ for the semi-supervised one. The algorithm is implemented as in Alg.~\ref{alg: power_iteration} with the multi-channels as in Eq.~\ref{eq: multi_channel}.

\settasks{
  label-offset = 0em ,
  item-indent = 0em ,
  item-indent = 1em ,
  column-sep = 1em
}

\setlength{\tabcolsep}{0.2em}
\begin{table}[h]
\begin{center}
	\begin{tabular}{l r c c c}
		\toprule
         &
        \multicolumn{1}{p{2cm}}{\raggedleft Input\\Method} &
        \multicolumn{1}{p{1cm}}{\centering Input Score (J)} &
        \multicolumn{1}{p{1.6cm}}{\centering \textbf{SFSeg}\\over\\Input (J)} &
        \multicolumn{1}{p{1.5cm}}{\centering Improved\\Videos\\ (\%)} \\
        \midrule
		Semi & OnAVOS  & 86.1 & \textbf{86.3} (+0.2)& 65 \\
		Supervised  & OSVOS-S & 85.6 & \textbf{86.0} (+0.4)& 90\\
	    & PReMVOS & 84.9 & \textbf{88.2} (+3.3)& 90\\
        & FAVOS  & 82.4 & \textbf{83.0} (+0.6)& 95 \\
        & OSMN  & 73.9 & \textbf{75.9} (+2.0) & 95 \\
        \midrule
		Un  & COSNet & 80.5 & \textbf{80.9} (+0.4) & 65 \\
		Supervised & MotAdapt & 77.2 & \textbf{77.5} (+0.3) & 65 \\
		& PDB  & 77.2 & \textbf{77.4} (+0.2) & 60\\
		& ARP & 76.2 & \textbf{77.7} (+1.5) & 90 \\
	    & LVO  & 75.9 & \textbf{78.8} (+2.9) & 90 \\
	    & FSEG & 70.7 & \textbf{72.3} (+1.6) & 95\\
	    & NLC  & 55.1 & \textbf{55.6} (+0.5) & 65\\
		\midrule
		& Average Boost  &  &  +1.1\% & 80\%\\
		\bottomrule
    \end{tabular}
\end{center}			
\caption{Improvement over top segmentation methods on DAVIS-2016 tasks, validation set. SFSeg has the same hyper-parameters per task. We also included results for other competitive (non-SOTA) inputs. $2^{nd}$ column: Jaccard score for the input method; $3^{rd}$ column: score after applying SFSeg over the input method; $4^{th}$ column: the percentage of videos when the performance is improved after using SFSeg. The average SFSeg boost is 1.1\% in Jaccard score and on average SFSeg raises performance for 80\% of videos. Input methods : \protect\cite{onavos,osvoss,premvos,cosnet,motadapt,pdb,arp,lvo,fseg,favos,osmn,nlc}.
}
\label{tab: davis_results}

\end{table}

\begin{figure}[!h]
	\begin{center}
		\includegraphics[width=0.85\linewidth]{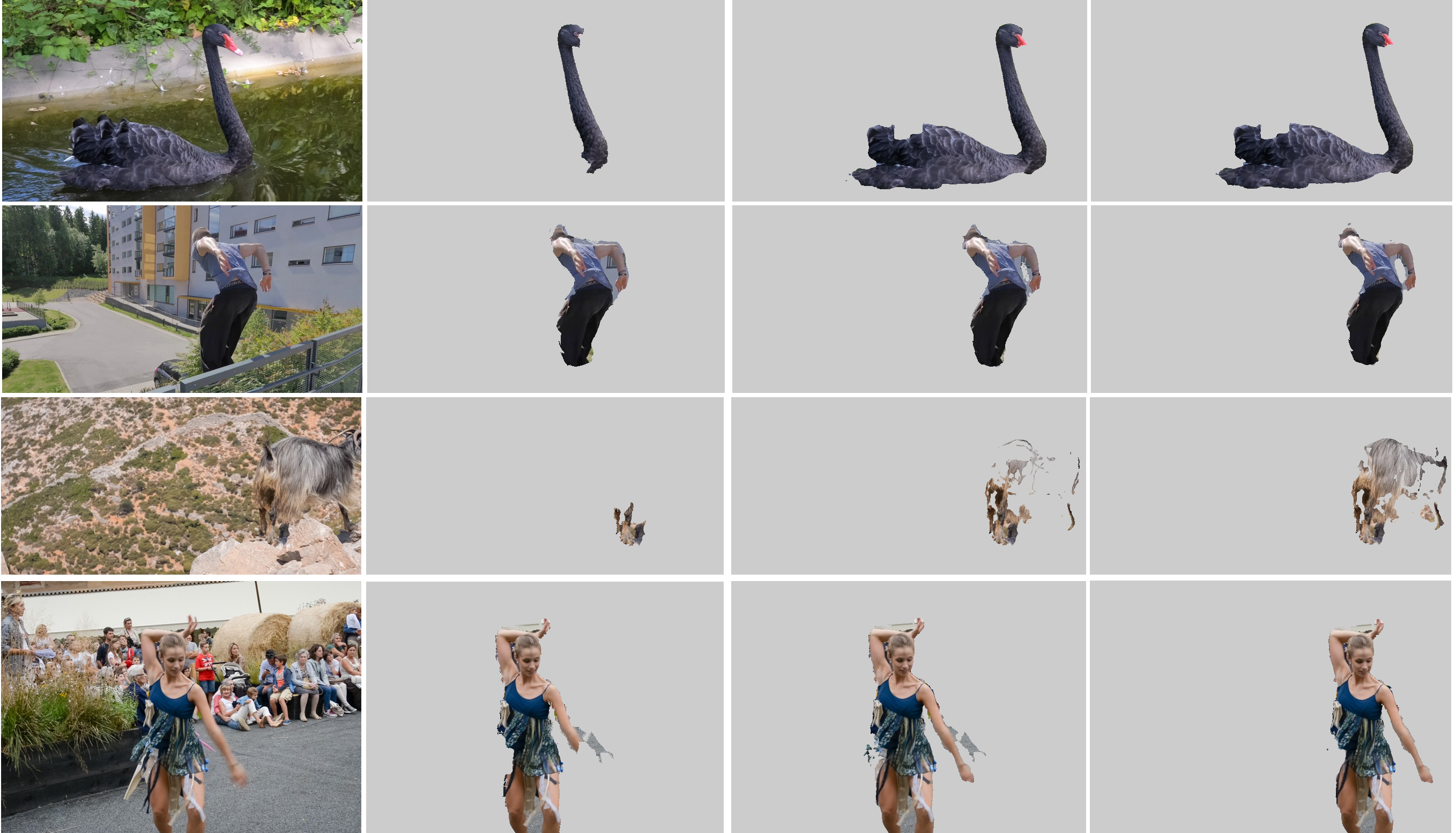}
		\vspace*{-2mm}
		\begin{tasks}(4)
          \task[Input Frames]
          \task[Input Mask]
          \task[SFSeg\\Iter2]
          \task[SFSeg\\Convergence]
        \end{tasks}
	\end{center}
	\caption{We present the evolution of SFSeg, over several iterations. Using the input segmentation mask (column 2) from top methods on DAVIS: ARP, FSEG and LVO, we show the intermediate value of the mask at Iter2 (column 3) and Iter4 (column 4).}
	\label{fig: qualitative_iterations}
\end{figure}

In Tab.~\ref{tab: davis_results} we show the results of our method, SFSeg, when combined with top methods on DAVIS-2016, semi-supervised and unsupervised tasks. For a better understanding of the results, we also show the effect of applying SFSeg over other competitive, non-SOTA methods. We noted that the improvement is not related with the quality measure of the input. In some cases the improvement is stronger when input comes from stronger methods. Nevertheless, we consistently improve over the input method, whose segmentation mask we use to initialize the segmentation $\mathbf{X}_0$.

\begin{figure*}[!t]
     \begin{minipage}[t]{0.32\linewidth}
     \raggedright
        \includegraphics[width=0.95\textwidth]{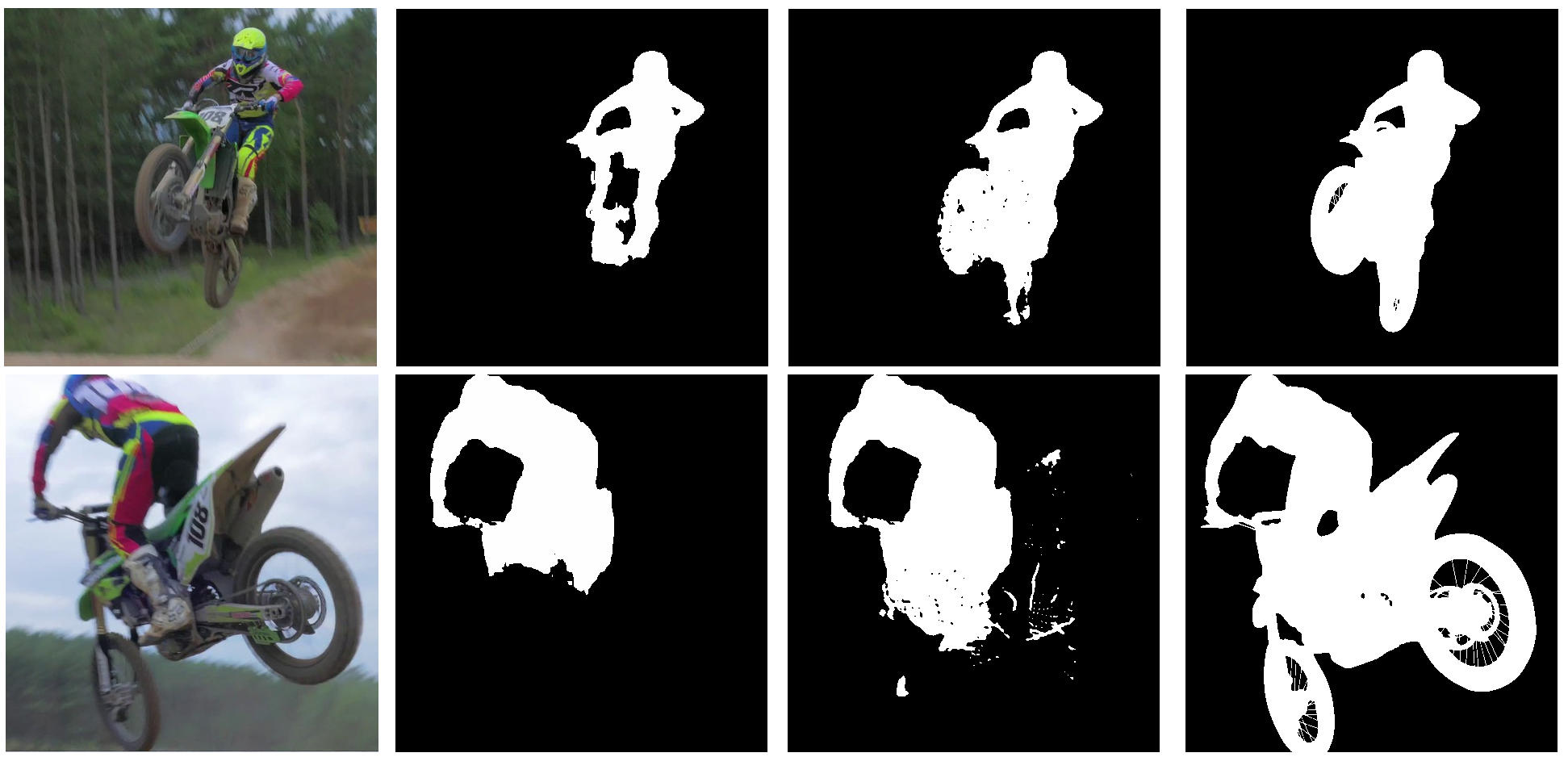}
        \vspace*{-2mm}
        \begin{tasks}(4)
          \task[Input]
          \task[\small PReMVOS]
          \task[SFSeg]
          \task[GT]
        \end{tasks}
	\end{minipage}
	\begin{minipage}[t]{0.32\linewidth}
	\centering
        \includegraphics[width=0.95\textwidth]{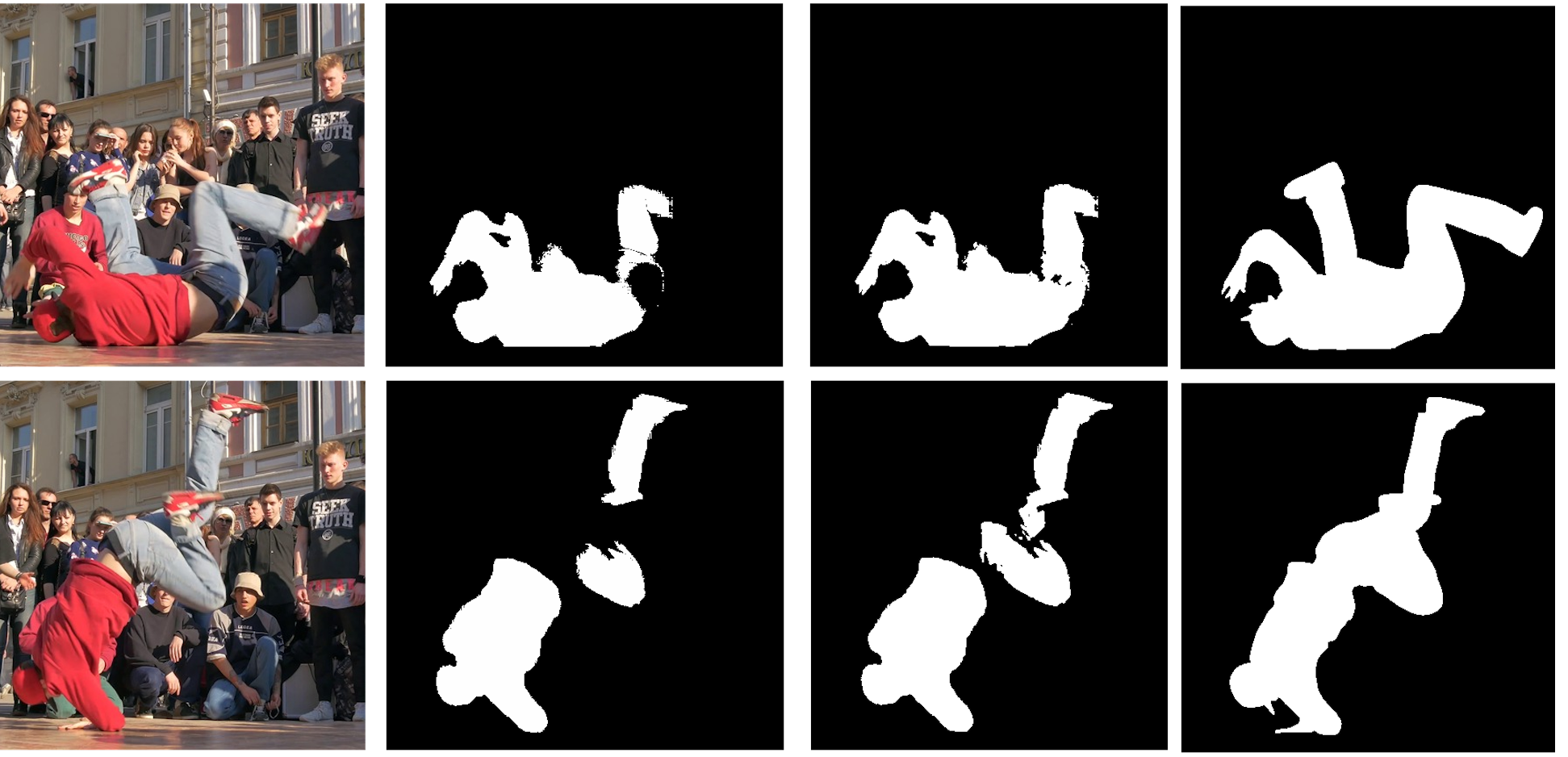}
        \vspace*{-2mm}
        \settasks{
          item-indent = 0.2em ,
        }
        \begin{tasks}(4)
          \task[Input]
          \task[\small OnAVOS]
          \task[SFSeg]
          \task[GT]
        \end{tasks}
	\end{minipage}
  	\begin{minipage}[t]{0.32\linewidth}
  	\raggedleft
         \includegraphics[width=0.95\textwidth]{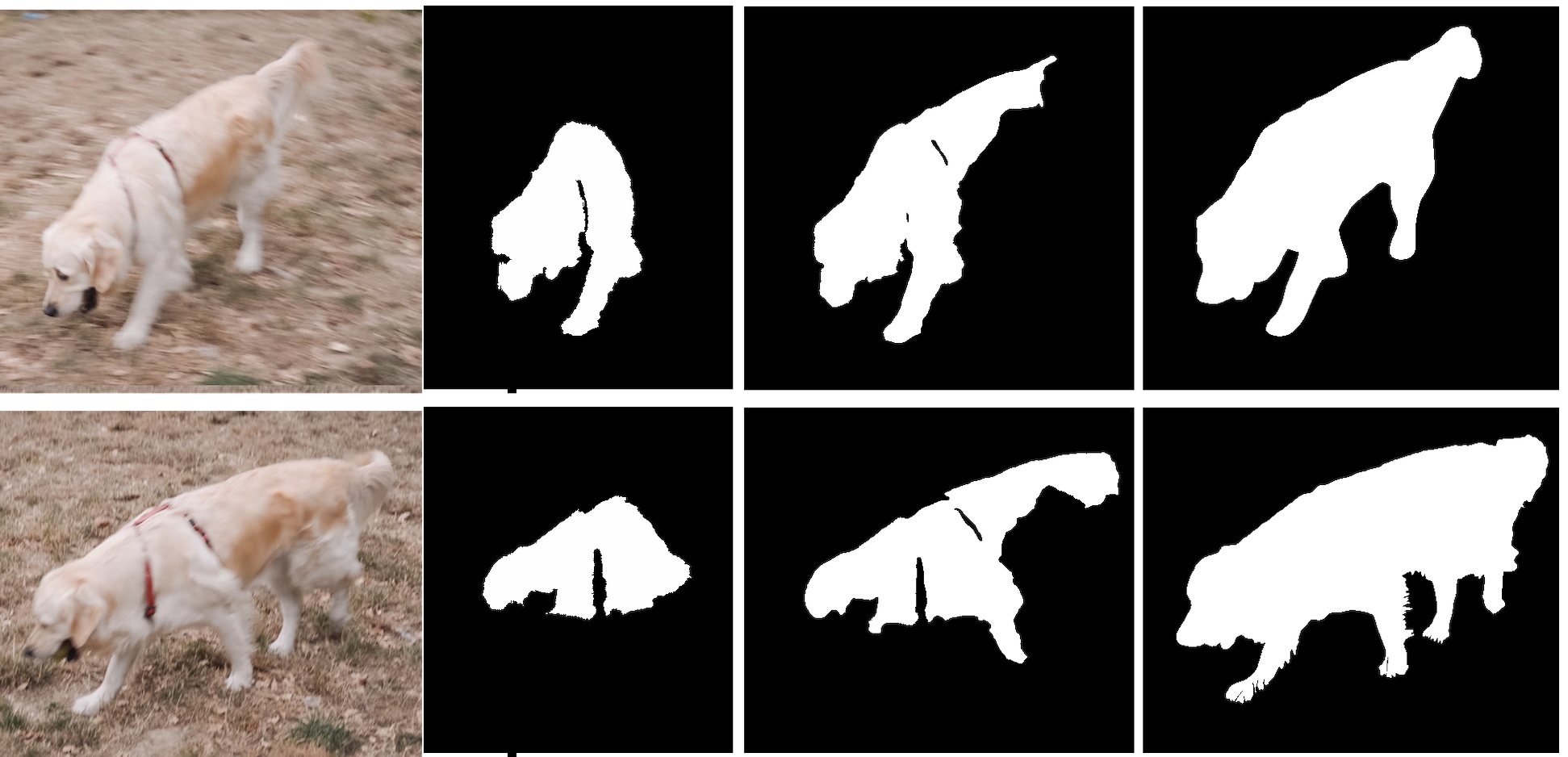}
        \vspace*{-2mm}
        \settasks{
          item-indent = 1.7em,
          column-sep = 0em
        }
        \begin{tasks}(4)
          \task[Input]
          \task[\small ARP]
          \task[SFSeg]
          \task[GT]
        \end{tasks}
	\end{minipage}
	\caption{We show the output of SFSeg (col 3) over the input masks (col 2) received from top DAVIS-2016 solutions in various video frames (col 1). We see how the quality of the masks is increasing, bringing the input masks closer to ground truth (col 4). 1. PReMVOS - $3^{rd}$ place on semi-supervised (motocross-jump); 2. OnAVOS   - $1^{st}$ place on semi-supervised (breakdance); 3. ARP - $2^{nd}$ place on unsupervised (dog).}
	\label{fig: qualitative_hard}
\end{figure*}
In Fig.~\ref{fig: qualitative_iterations} we show the iterative effect of SFSeg. Each example starts from the initial RGB frame and its initial segmentation (as produced by top DAVIS methods), and presents the segmentation at an intermediate iteration and the final one, when SFSeg reaches convergence.

In Fig.~\ref{fig: qualitative_hard} we show qualitative examples of our spectral method, SFSeg, applied over the initial mask, received as input from highly qualitative segmentation solutions on DAVIS-2016. The new masks show significant improvement without using other new means of supervision.

\paragraph{Experiments on SegTrackv2.} SegTrackv2~\cite{segtrack2} is a video object segmentation dataset, containing 14 videos, with multiple objects per frame. The purpose for video object segmentation task is to find the segmentation for all the objects in the frame (also split in two tasks: using the first frame or not). We use our standalone method, SFSeg, applied over the soft output of a competitive \textbf{Backbone (BB)}: UNet over ResNet34 pretrained features, fine tuned 40 epochs on salient object segmentation in images on DUTS dataset \cite{saliencyDS} (with RectifiedAdam as optimizer). In Tab.~\ref{tab:segtrack2} we show comparative results of our standalone method and other top solutions on the SegTrackv2 dataset.

\begin{table}[t]
\begin{center}
	\begin{tabular}{r r r r}
		\toprule
		\multicolumn{1}{r}{Method} &
        \multicolumn{1}{r}{Score (J)}\\
        \midrule
        LVO & 57.3\\
        FSEG & 61.4\\
		OSVOS & 65.4\\
		NLC  & 67.2\\
		MaskTrack & 70.3\\
		\textbf{BB} + \textbf{SFSeg} + denseCRF (ours)& \textbf{72.7} \\
        \bottomrule
    \end{tabular}
\end{center}			
\caption{Comparative results on SegTrackv2. Our standalone solution, Backbone + SFSeg + denseCRF, obtains the best results among the other top methods in the literature. Input methods: \protect\cite{lvo,fseg,osvos,nlc,masktrack}.}
\label{tab:segtrack2}
\end{table}

\paragraph{SFSeg vs denseCRF.} We compare SFSeg with denseCRF \cite{denseCRF}, which is one of the most used refinement method in video object segmentation \cite{pdb}. When applied over the same \textbf{Backbone} presented above, we observe that SFSeg brings a stronger improvement than denseCRF on both DAVIS-2016 and SegTrackv2 (Tab.~\ref{tab:backbone}). More, the two are complementary: in combination, the performance is boosted by the largest margin.

\begin{table}[t]
\begin{center}
	\begin{tabular}{l r r r}
		\toprule
		\multicolumn{1}{l}{Method} &
		\multicolumn{1}{p{1cm}}{\raggedleft DAVIS (J)} &
		\multicolumn{1}{p{2.1cm}}{\raggedleft SegTrackv2 (J)} \\
        \midrule
		\textbf{BB} & 67.2 & 72\\
		\textbf{BB} + denseCRF & 68.1 & 72\\
		\textbf{BB} + SFSeg& 68.7 & 72.1 \\
		\textbf{BB} + SFSeg + denseCRF & \textbf{69.2} & \textbf{72.7} \\
		\bottomrule
    \end{tabular}
\end{center}
\caption{Refinement Comparison. We compare SFSeg with denseCRF when applied to a competitive end-to-end Backbone (as detailed above). While SFSeg outperforms denseCRF when used individually, the two methods prove to be not only different, but also complementary, since combining them boosts the Jaccard score.}
\label{tab:backbone}
\vspace{-1em}
\end{table}

\paragraph{Running Time.} The algorithm scales well, the running time being linear in the number of video pixels, as detailed in Sec.~\ref{sec: algorithm}. For a frame of $480\times854$ pixels, it takes $0.055$ sec per iteration, compared with $0.8$ sec for denseCRF. Filtering takes 60\% of time, the rest of 40\% being used on other auxiliary operations (copying tensors). The time penalty of adding SFSeg is minor for most methods, which takes several seconds per frame (\eg $4.5$ sec per frame \cite{osvoss}, $13$ sec per frame \cite{premvos}). We tested on a GTX Titan X Maxwell GPU, in Pytorch \cite{pytorch}.

\section{Conclusions and Future Work}
\label{sec: conclusions}
We formulate video object segmentation as clustering in the space-time graph of pixels.
We introduce an efficient spectral algorithm, Spectral Filtering Segmentation (SFSeg), in which the standard power iteration for computing the principal eigenvector of the graph adjacency matrix is transformed into a set of 3D convolutions applied on 3D feature maps in the video volume. Our original theoretical contribution makes the initial intractable problem possible. We validate experimentally that our solution based on a first-order Taylor approximation of the original pairwise potential used in spectral clustering is practically equivalent to the original one. In experiments, SFSeg consistently improves (for 80\% of videos) over top published video object segmentation methods, at a small additional computational cost. Moreover, our method also achieves top performance in combination with other backbone networks (not necessarily state of the art). The consistent improvements in practice indicate that our spectral approach brings a new and complementary dimension to clustering in space-time, which is not fully addressed by current solutions. In the immediate future we will explore ways to learn more powerful features end-to-end, within our spectral clustering formulation.

\clearpage 
\bibliographystyle{named}
\bibliography{ijcai20}

\end{document}